\documentclass[letterpaper, 10 pt, conference]{ieeeconf}  

\IEEEoverridecommandlockouts 
\usepackage{placeins}
\usepackage{graphicx}
\usepackage{cite}
\usepackage{amsmath,amssymb,amsfonts}
\usepackage{graphicx}
\usepackage{textcomp}
\usepackage{xcolor}
\usepackage{caption}
\usepackage[font=small]{subcaption}
\usepackage{multirow}
\usepackage{comment}
\usepackage[]{todonotes}
\usepackage{longtable,tabu}
\usepackage{layouts}
\usepackage{tabularx}

\usepackage{comment}
\usepackage[linesnumbered,ruled]{algorithm2e}
\usepackage[noend]{algpseudocode}
\usepackage{textgreek}
\usepackage[shortcuts]{extdash}
\usepackage{multirow}
\usepackage{graphicx}
\usepackage{multirow}
\usepackage{graphicx}
\usepackage{rotating}
\usepackage{amsmath}
\usepackage{hyperref}
\newcommand*{\vsepfbox}[1]{%
  \begingroup
    \sbox0{\fbox{#1}}%
    \setlength{\fboxrule}{0pt}%
    \mbox{\kern-\fboxsep\fbox{\unhbox0}\kern-\fboxsep}%
  \endgroup
}

\title{\LARGE \bf
Runtime Detection of Executional Errors in Robot-Assisted Surgery 
}

\author{Zongyu Li, Kay Hutchinson, and Homa Alemzadeh
\thanks{*This work was partially supported by an award from the Engineering-in-Medicine center at the University of Virginia and the NSF NRT program (Grant No. 1829004).}
\thanks{Authors are with the Department of Electrical and Computer Engineering (ECE), University of Virginia, Charlottesville, VA 22904, USA.
        {\tt\small \{zl7qw, kch4fk, ha4d\}@virginia.edu}}%
\thanks{The corresponding code for this paper is available at \url{https://github.com/UVA-DSA/ExecErr_Detect_Siamese}}
}

\begin{document}

\maketitle
\thispagestyle{empty}
\pagestyle{empty}
\begin{abstract}
Despite significant developments in the design of surgical robots and automated techniques for objective evaluation of surgical skills, there are still challenges in ensuring safety in robot-assisted minimally-invasive surgery (RMIS). This paper presents a runtime monitoring system for the detection of executional errors during surgical tasks through the analysis of kinematic data. The proposed system incorporates dual Siamese neural networks and knowledge of surgical context, including surgical tasks and gestures, their distributional similarities, and common error modes, to learn the differences between normal and erroneous surgical trajectories from small training datasets. We evaluate the performance of the error detection using Siamese networks compared to single CNN and LSTM networks trained with different levels of contextual knowledge and training data, using the dry-lab demonstrations of the Suturing and Needle Passing tasks from the JIGSAWS dataset. Our results show that gesture specific task nonspecific Siamese networks obtain micro F1 scores of 0.94 (Siamese-CNN) and 0.95 (Siamese-LSTM), and perform better than single CNN (0.86) and LSTM (0.87) networks. These Siamese networks also outperform gesture nonspecific task specific Siamese-CNN and Siamese-LSTM models for Suturing and Needle Passing.

\end{abstract}

\section{INTRODUCTION}






Robot-assisted minimally invasive surgery (RMIS) has become a standard approach across different specialties, including urology, gynecology, and general surgery. 
Surgical robots translate the surgeon's hand, wrist, and finger movements into precise motions of robotic manipulators and miniaturized surgical instruments. 
Potential benefits include enhanced visualization, increased dexterity, smaller incisions, and shorter recovery time. 
However, the safety of RMIS can be compromised due to unintentional human errors \cite{hutchinson2021analysis, elhage2015assessment, eubanks1999objective, joice1998errors} or the potential vulnerabilities of surgical robots to accidental faults and malicious attacks targeting sensors, actuators~\cite{alemzadeh2016adverse}, communication between the surgeon's console and the robot~\cite{bonaci2015make, bonaci2015experimental}, or the robot control software \cite{Alemzadeh_model_2016, alemzadeh2014systems}. 


Developing machine learning (ML) techniques for automated and objective evaluation of surgical skills based on kinematic \cite{Wang_Majewicz_Fey_2018,Zia_skill_hand} and video \cite{Funke_Mees_Weitz_Speidel_2019} data from procedures has been an active area of research. However, the current ML techniques are mostly developed for \textit{offline} analysis of surgeon's performance. The next generation of surgical robots and simulators can be enhanced with capabilities for \textit{runtime} analysis of surgical tasks and providing data-driven feedback to surgeons during training or actual surgery to improve safety, efficiency, and quality of care~\cite{maier2017surgical, Yasar_Alemzadeh_2020}. 

Previous works have proposed the idea of runtime surgical monitoring for detection and prevention of safety-critical events~\cite{Alemzadeh_model_2016,Yasar_context_2019,Yasar_Alemzadeh_2020}, surgeon authentication~\cite{yan2021continuous}, and virtual coaching~\cite{maier2017surgical}. In \cite{Alemzadeh_model_2016}, the authors developed a dynamic model-based anomaly detection technique for mitigation of unsafe events caused by malicious attacks on a surgical robot controller. However, errors can also happen in the operational context as a result of surgeons' sub-optimal performance~\cite{Yasar_context_2019}. In \cite{hutchinson2021analysis}, a new rubric was proposed for evaluating executional and procedural errors based on video and kinematic data from dry-lab simulation experiments. This work showed that the type and frequency of errors are dependent on surgical context, characterized by the specific tasks (e.g., Suturing) and gestures (e.g., Pulling suture) performed by the surgeon. 

Recently, deep learning (DL) techniques (e.g., CNN and LSTM networks), have been applied to the prediction of potential unsafe events caused by unintentional human errors in simulated surgical training tasks~\cite{Yasar_context_2019,Yasar_Alemzadeh_2020} and retinal microsurgery~\cite{He_safety_2019}. These works showed preliminary evidence for the importance of incorporating the knowledge of surgical gestures in model training ~\cite{Yasar_context_2019,Yasar_Alemzadeh_2020} to improve detection performance. However, the performance of DL models is often negatively affected by the small size of training data and the lack of labeled datasets on errors~\cite{Yasar_Alemzadeh_2020}.


In this paper we aim to address the challenges in runtime detection of executional errors in robot-assisted surgery by developing models that can achieve reliable performance given small training data. Our goal is to investigate whether learning the differences among erroneous and normal trajectories, and incorporating different levels of surgical context information can improve the error detection accuracy. The main contributions of the paper are as follows:

\begin{itemize}
    \item Developing a runtime monitoring system based on dual Siamese neural network architectures that learns the differences between normal and erroneous kinematic trajectories and enables reliable error detection performance using small training dataset sizes.
    \item Demonstrating that incorporating the knowledge of surgical context, including information on the current surgical task and gesture, which captures distributional similarities across gesture and task trajectories and specific types and frequencies of errors in model training, can improve the error detection performance. 
    \item Evaluating the proposed system for runtime detection of executional errors using the publicly-available JIGSAWS dataset collected from dry-lab robot-assisted surgical tasks (Suturing and Needle Passing) performed on a da Vinci robot from Intuitive Surgical Inc. \cite{daVinci}. Our experiments show that dual input Siamese neural networks can achieve on average a better performance than single CNN and LSTM models after augmenting the data with more trajectories (e.g., for gesture nonspecific and task specific models, F1 score for Siamese-CNN vs. CNN: 0.79 vs. 0.72 and Siamese-LSTM vs. LSTM: 0.87 vs. 0.60). We also found that incorporating the knowledge of surgical gestures and specifically training the models with gesture specific trajectory examples helps with improving F1 scores more than training with more gesture nonspecific trajectory examples.
    
\end{itemize}

\section{Problem Statement}
\begin{figure*}[t!]
    \centering
    \includegraphics[width=0.85\textwidth]{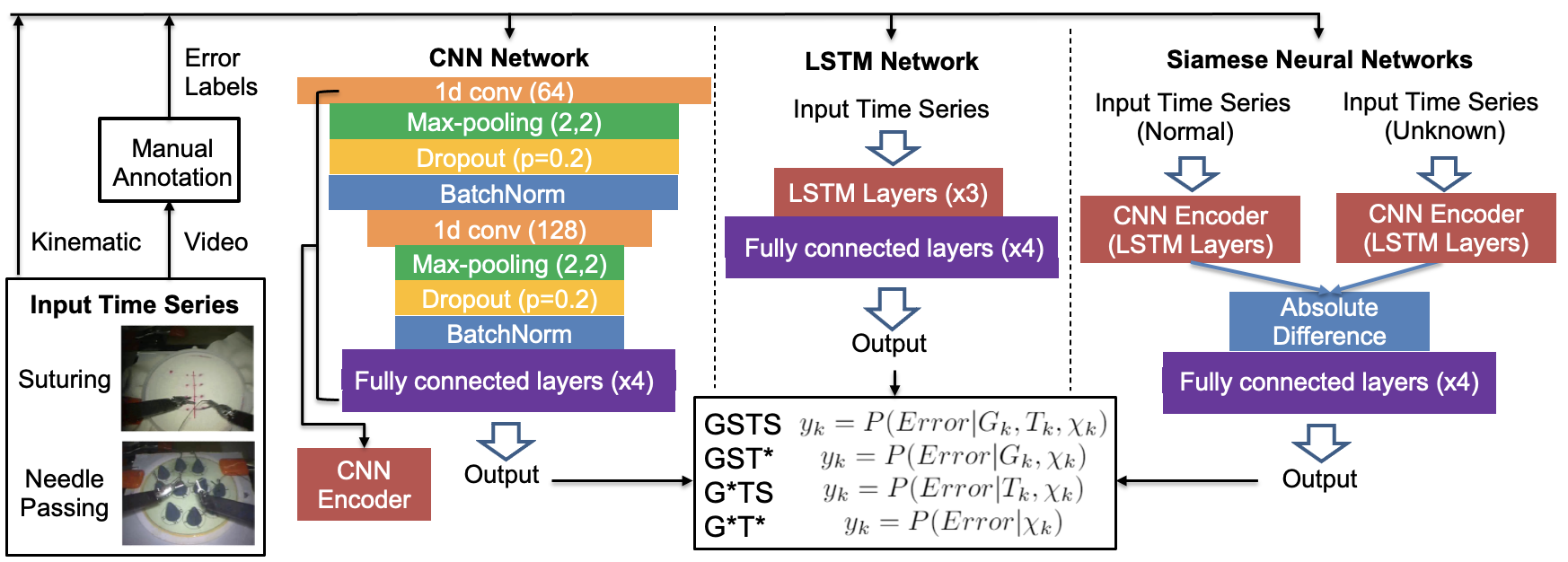}
    \vspace{-0.5em}
    \caption{Different Neural Network Structures with Different Training Setups and Contextual Information}
    \vspace{-1.5em}
    \label{fig:diff_networks}
\end{figure*}

We model surgical procedures as a hierarchy composed of a sequence of tasks (e.g., Suturing, Needle Passing) coming from a library of surgical tasks $T$, each represented by a grammar graph with steps defined as gestures which are the atomic surgical activities in a procedure~\cite{Yasar_Alemzadeh_2020, hutchinson2021analysis}. We denote the library of all surgical gestures with $G$ such as in Table \ref{tab:gesture_error}. The change in the surgical context happens in the temporal domain as a result of the change in the position and orientation of the surgical instruments' end-effectors, leading to the corresponding change in the gestures and tasks. 

Our previous analysis of data from dry-lab robotic surgery experiments showed that executional and procedural errors can negatively affect surgical performance~\cite{hutchinson2021analysis}. Executional errors are defined as the failure of a specific step in a task. Procedural errors are defined as the omission or re-arrangement of correctly undertaken steps within a task \cite{joice1998errors}. Our goal is to detect executional errors at runtime by analyzing the kinematic data collected from the robot. 

The input kinematic data to our error detection system is a multi-dimensional time series matrix $\chi_{k} \in \mathbb{R}^{N_{ch} \times N_s}$, 
with $N_{ch}$ representing the number of features extracted from the robot's kinematic data and $N_{s}$ representing the size of the sliding window. We model the data pre-processing step as a function $f(\cdot)$ on each input window $\chi_k$. 
We assume each $\chi_k$ can be mapped to a gesture class $G_{i} \in G$ using a gesture segmentation and classification method like those proposed in the literature~\cite{Gao2020gestures,Amsterdam2020MultiTaskRN,Amsterdam2019gesture,Qin2020gesture,Hager2016TCN,Hager2016RNN,Goldberg2017unsupervised,Tao2012HMM,Varadarajan2009seg}. We adopt the gesture-specific error rubric presented in \cite{hutchinson2021analysis} to define a set of unique executional error modes $E$ corresponding to each gesture class as shown in Table \ref{tab:gesture_error}. The output of the error detector is an error probability $y_{k}$ assigned to the input window $\chi_k$, as shown in Figure \ref{fig:diff_networks}, where $G_{k} $ and $T_{k}$ represent the gesture and task classes that the input is mapped to and $Error$ represents the event that one of the error modes in $E$ is observed in this window.


Our goal is to investigate the following research questions by designing and evaluating different neural network architectures for runtime error detection using different levels of contextual information and amounts of training data: 
   
   \textbf{RQ1:} Can learning the patterns of differences between erroneous and normal trajectories using a dual input network rather than just learning error patterns improve error detection performance? 
   
   \textbf{RQ2:} How does incorporating the knowledge of the tasks and gestures being performed by the surgeon/robot affect the performance of erroneous gesture detection?
   
   We specifically consider the following setups for training different models with different levels of contextual knowledge on the tasks and gestures:
   \begin{itemize}
        \item Gesture specific and task specific models
        \item Gesture specific and task nonspecific models
        \item Gesture nonspecific and task specific models
        \item Gesture nonspecific and task nonspecific model
    \end{itemize}
    We hypothesize that the models trained with the data specific to a task or gesture can achieve better performance than the models with no contextual knowledge. 


\begin{table}[t]
\vspace{0.15cm}
    \centering
    
    \caption{Gesture Definitions (adopted from \cite{JIGSAWS}), Common Gesture-specific Executional Error Modes (adopted from \cite{hutchinson2021analysis})}
    \vspace{-0.5em}
    \resizebox{\linewidth}{!}{
    \begin{tabular}{|p{0.1\linewidth}|p{0.4\linewidth}|p{0.25\linewidth}|p{0.2\linewidth}|}
\hline
\textbf{Gesture} & \textbf{Description}                           & \textbf{Common Gesture-Specific Errors} \\ \hline
G1       & Reaching for needle with right hand     & Multiple attempts               \\ \hline
G2       & Positioning needle                      & Multiple attempts, Out of view  \\ \hline
G3       & Pushing needle through the tissue       & Multiple attempts               \\ \hline
G4 & Transferring needle from left to right                    & Multiple attempts, Needle orientation \\ \hline
G6       & Pulling suture with left hand           & Multiple attempts, Out of view  \\ \hline
G8 & Orienting needle \& Uses tissue/ instrument for stability & Multiple attempts, Needle orientation \\ \hline
G9       & Using right hand to help tighten suture & Multiple attempts               \\ \hline
\end{tabular}%
}
\vspace{-2em}
\label{tab:gesture_error}
\end{table}

\section{Methods} 
\label{sec: Methods}
We develop different single and dual neural network architectures for runtime detection of executional errors, including a CNN network and an LSTM network (inspired by state-of-the-art gesture recognition and error detection models \cite{Yasar_Alemzadeh_2020,Hager2016TCN}) and their corresponding Siamese networks, respectively built with CNN encoders and LSTM layers (see Figure \ref{fig:diff_networks}). This section describes the structure of these networks, the data pre-processing done for training and testing the networks, and how the output from the models were utilized to detect erroneous gestures.

\subsection{Data Pre-processing}
The input to the error detection networks is a 26-dimensional time-series matrix. To reduce the computational cost and enable faster training convergence, we time-normalized and downsampled the input time-series with a factor of 2. A sliding window of 30 samples (derived from the median length of the shortest gesture in the dataset) with a stride of 20 samples was used to obtain the input data.

\subsection{CNN Network}
The CNN network shown in Figure \ref{fig:diff_networks} is designed with two sets of blocks composed of a 1d convolution layer, a max-pooling layer, a dropout layer, and a batch normalization layer. The output from the last batch normalization layer is flattened and fed into four fully connected layers, including three layers with ReLU activation functions \cite{nair2010rectified} and one last layer with a Sigmoid activation function that outputs the probability of the gesture being erroneous. To train the network, $[f(\chi_k),Y_{k}]$ is fed into the network, where $f(\chi_k)$ is the data after pre-processing, and $Y_{k}$ is the ground truth indicating whether the gesture is erroneous ($Y_{k}=1$) or normal $(Y_{k}=0)$. The binary cross entropy loss function \cite{crossentropy} is used to train this network.

\subsection{LSTM Network}
The LSTM network contains 3 LSTM layers. The hidden states from the last LSTM layer are flattened and fed into four fully connected layers with dropout layers and  with ReLU activation functions for the first three layers. The last fully connected layer is connected to a Sigmoid activation function that outputs the probability of a gesture being normal or erroneous. The training process is similar to the CNN network.

\subsection{Dual Siamese Neural Networks}

\par Siamese neural networks are a class of neural network architectures with two or more identical structures that focus on learning a similarity function for indicating whether the inputs are from the same or different classes. Because of their ability to achieve better prediction accuracy than single networks for small data sizes, they have been applied to applications such as signature verification, face recognition~\cite{Koch_2015Siamese}, and EEG-based brain-computer interface signal classification~\cite{Shahtalebi_Asif_Mohammadi_2020}.

As shown in Figure \ref{fig:diff_networks}, we designed two different Siamese networks, each containing two identical neural networks, with the same structure as the single input CNN encoder or the LSTM layers used for the CNN and LSTM networks described above. For binary classification, each Siamese network takes in two gesture windows and processes them using its two identical structures in parallel. Then, the absolute value of the difference between outputs from the two structures is calculated. We then use three fully connected layers with ReLU activation, and the fourth layer is connected to a Sigmoid activation function that outputs the probability that one of the two windows belongs to a different class. 

\par  To train the Siamese neural network, the tuple $[f(\chi_{k}),f(\chi_{j}),Y_{kj}]$ is fed into the network, where $f(\chi_{k}),f(\chi_{j})$ are pairs of the gesture windows after pre-processing and $Y_{kj}$ denotes the label. $Y_{kj}$ is assigned a value of 0 when $\chi_{k},\chi_{j}$ are both normal gesture windows and a value of 1 when one of the windows is erroneous and the other is normal. The binary cross entropy loss function is used to train the network.

During the evaluation phase, each input window is paired with all the existing normal windows in the training dataset. The network outputs a value of 0 when the input window is indicated to be in the same class as a normal window and a value of 1, meaning that the unknown gesture window is not normal, and thus, erroneous. We then use majority voting to fuse the results from all the pairs to generate a final normal or erroneous predicted label for each input window. 

\begin{table*}[t!]
\vspace{0.13cm}
\resizebox{\textwidth}{!}{%
\begin{tabular}{|l|l|l|l|l|l|l|l|l|lll}
\cline{1-9} \cline{11-12}
\textbf{Training}                  & \textbf{Task}                                                 & \textbf{Network}          & \textbf{G1}            & \textbf{G2}            & \textbf{G3  }          & \textbf{G4  }          & \textbf{G6}            & \textbf{micro F1}   & \multicolumn{1}{l|}{} & \multicolumn{1}{c|}{\textbf{micro F1}}   & \multicolumn{1}{l|}{\textbf{Training}}                 \\ \cline{1-9} \cline{11-12} 
\multirow{12}{*}{\begin{turn} {90}\begin{tabular}[c]{@{}l@{}}\textbf{Gesture specific}\\\textbf{Task specifiC}\\\textbf{(GSTS)}\end{tabular} \end{turn}} & \multicolumn{1}{l|}{\multirow{4}{*}{Suturing}}       & CNN              & \textbf{0.66} & \textbf{0.48} & \textbf{0.73} & 0.72          & 0.90          & \textbf{0.78}       & \multicolumn{1}{l|}{} & \multicolumn{1}{l|}{0.72}       & \multicolumn{1}{l|}{\multirow{12}{*}{\begin{turn} {90}\begin{tabular}[c]{@{}l@{}}\textbf{Gesture nonspecific}\\\textbf{Task specific}\\\textbf{(G*TS)}\end{tabular}\end{turn}}} \\ \cline{3-9} \cline{11-11}
                       & \multicolumn{1}{c|}{}                                & Siamese -CNN     & 0.58          & 0.29          & 0.70          & \textbf{0.75} & \textbf{0.91}          & 0.58       & \multicolumn{1}{l|}{} & \multicolumn{1}{l|}{\textbf{0.79}}       & \multicolumn{1}{l|}{}                      \\ \cline{3-9} \cline{11-11}
                       & \multicolumn{1}{c|}{}                                & LSTM             & 0.52          & \textbf{0.48} & 0.73          & \textbf{0.84} & \textbf{0.94} & 0.52       & \multicolumn{1}{l|}{} & \multicolumn{1}{l|}{0.60}       & \multicolumn{1}{l|}{}                      \\ \cline{3-9} \cline{11-11}
                       & \multicolumn{1}{c|}{}                                & Siamese -LSTM    & \textbf{0.66} & 0.33          & \textbf{0.74} & 0.75          & 0.90          & \textbf{0.66}       & \multicolumn{1}{l|}{} & \multicolumn{1}{l|}{\textbf{0.87}}       & \multicolumn{1}{l|}{}                      \\ \cline{2-9} \cline{11-11}
                       & \multirow{2}{*}{Data}                                & Total Trials     & 29            & 166           & 164           & 119           & 163           & 641        & \multicolumn{1}{l|}{} & \multicolumn{1}{l|}{641}        & \multicolumn{1}{l|}{}                      \\ \cline{3-9} \cline{11-11}
                       &                                                      & Erroneous Trials & 8 (28\%)      & 22 (13\%)     & 82 (51\%)     & 71 (60\%)     & 121 (74\%)    & 304 (47\%) & \multicolumn{1}{l|}{} & \multicolumn{1}{l|}{304 (47\%)} & \multicolumn{1}{l|}{}                      \\ \cline{2-9} \cline{11-11}
                       & \multicolumn{1}{c|}{\multirow{4}{*}{Needle Passing}} & CNN              & \textbf{0.57} & \textbf{0.61} & 0.36          & \textbf{0.30} & 0.43          & 0.43       & \multicolumn{1}{l|}{} & \multicolumn{1}{l|}{0.56}       & \multicolumn{1}{l|}{}                      \\ \cline{3-9} \cline{11-11}
                       & \multicolumn{1}{c|}{}                                & Siamese -CNN     & 0.51          & 0.55          & \textbf{0.43} & 0.20          & \textbf{0.46} & 0.43       & \multicolumn{1}{l|}{} & \multicolumn{1}{l|}{\textbf{0.74}}       & \multicolumn{1}{l|}{}                      \\ \cline{3-9} \cline{11-11}
                       & \multicolumn{1}{c|}{}                                & LSTM             & 0.52          & \textbf{0.82} & \textbf{0.75} & \textbf{0.42} & 0.51          & \textbf{0.75}       & \multicolumn{1}{l|}{} & \multicolumn{1}{l|}{0.56}       & \multicolumn{1}{l|}{}                      \\ \cline{3-9} \cline{11-11}
                       & \multicolumn{1}{c|}{}                                & Siamese -LSTM    & \textbf{0.60} & 0.64          & 0.48          & 0.24          & \textbf{0.54} & 0.47       & \multicolumn{1}{l|}{} & \multicolumn{1}{l|}{\textbf{0.81}}       & \multicolumn{1}{l|}{}                      \\ \cline{2-9} \cline{11-11}
                       & \multirow{2}{*}{Data}                                & Total Trials  & 30            & 117           & 111           & 83            & 112           & 453        & \multicolumn{1}{l|}{} & \multicolumn{1}{l|}{453}        & \multicolumn{1}{l|}{}                      \\ \cline{3-9} \cline{11-11}
                       &                                                      & Erroneous Trials   & 11 (37\%)     & 55 (47\%)     & 17 (15\%)     & 23 (28\%)     & 46 (41\%)     & 152 (34\%) & \multicolumn{1}{l|}{} & \multicolumn{1}{l|}{152 (34\%)} & \multicolumn{1}{l|}{}                      \\ \cline{1-9} \cline{11-12} 
\multirow{6}{*}{\begin{turn} {90}\begin{tabular}[c]{@{}l@{}}\textbf{Gesture specific}\\\textbf{Task nonspecific}\\\textbf{(GST*)}\end{tabular}\end{turn}}   & \multirow{4}{*}{Both Tasks}                          & CNN              & 0.74          & 0.73          & 0.82          & 0.86          & 0.86          & 0.86       &                       &                                 &                                            \\ \cline{3-9}                                                                     &                                                      & Siamese -CNN                      & \textbf{0.89}                       & \textbf{0.87}                       & \textbf{0.92}                       & \textbf{0.87}                       & \textbf{0.94}                        & \textbf{0.94}                       &                       &                                 &                                                                                                                             \\ \cline{3-9} \cline{11-12} 
                                                                     &                                                      & LSTM                              & 0.72                       & 0.62                       & 0.83                       & 0.73                       & 0.87                        & 0.87                        & \multicolumn{1}{l|}{} & \multicolumn{1}{l|}{0.67}       & \multicolumn{1}{l|}{\multirow{2}{*}{\begin{tabular}[c]{@{}l@{}}\textbf{Baseline}\\ \textbf{G*T*}\end{tabular}}} \\ \cline{3-9} \cline{11-11}
                                                                     &                                                      & Siamese -LSTM                     & \textbf{0.89}                      & \textbf{0.87}                      & \textbf{0.93}                    & \textbf{0.90}                     & \textbf{0.95}                      & \textbf{0.95}                        & \multicolumn{1}{l|}{} & \multicolumn{1}{l|}{0.68}       & \multicolumn{1}{l|}{}                                                                                                       \\ \cline{2-9} \cline{11-12} 
                                                                     & \multirow{3}{*}{Data}                                & Total Trials                      & 59                         & 283                        & 275                        & 202                        & 275                         & 1094                        &                       &                                 &                                                                                                                             \\ \cline{3-9}
                                                                     &                                                      & \multirow{2}{*}{Erroneous Trials} & \multirow{2}{*}{19 (32\%)} & \multirow{2}{*}{77 (27\%)} & \multirow{2}{*}{99 (36\%)} & \multirow{2}{*}{94 (47\%)} & \multirow{2}{*}{167 (61\%)} & \multirow{2}{*}{456 (43\%)} &                       &                                 &                                                                                                                             \\
                                                                     &                                                      &                                   &                            &                            &                            &                            &                             &                             &                       &                                 &                                                                                                                             \\ \cline{1-9}
\end{tabular}%
\vspace{-1em}
}

\caption{Performance of CNN, Siamese-CNN, LSTM, and Siamese-LSTM Models with different Training Setups. Results for gestures G8 and G9 are not shown because of very small data available for these gestures (e.g., less than 3 erroneous trials in Needle Passing). We obtained the baseline \textbf{G*T*} models with our top performing network pairs (Siamese-LSTM/LSTM).} 
\label{tb:gesture_specific}
\vspace{-1.5em}
\end{table*}

\section{Experimental Evaluation}
\label{sec: experiment}

All experiments were conducted on a 64-bit PC with an Intel Core i9 CPU @  3.70GHz  and  32GB  RAM running Linux  Ubuntu  20.04 and an NVIDIA RTX2080 Ti 11GB GPU. The neural network models were implemented using PyTorch \cite{paszke2019pytorch} version 1.8.0.

\subsection{Datasets}
The JHU-ISI Gesture and Skill Assessment Working Set (JIGSAWS) \cite{JIGSAWS} is a publicly available dataset, collected using the da Vinci surgical robot \cite{daVinci} from eight  surgeons  with  different  levels  of  skills  and  expertise during three dry-lab surgical tasks (Suturing, Knot Tying, and Needle Passing) on a bench-top model. The JIGSAWS dataset includes synchronized kinematic and video data recorded at 30Hz along with manual annotations indicating the gestures for each trajectory and scores quantifying the surgeons' performance.
\par We used the kinematic data from the robot's patient-side  manipulators (PSM) for the Suturing and Needle Passing tasks and the gesture labels from the JIGSAWS dataset. The kinematic variables included tool tip position, orientation, linear velocity, rotational velocity, and gripper angle for the PSMs of both the left and right hands. Tool tip orientation was described using Euler angles calculated from the left and right hand rotational matrices to reduce the dimensionality of the inputs. 
We used the executional errors labels for JIGSAWS presented in \cite{hutchinson2021analysis}.  Each label for a gesture trial is either normal (0) or erroneous (1). Table \ref{tab:gesture_error}  summarizes the gesture definitions and their dominant error modes.
\subsection{Experimental Setup} 
We experimented with the following training setups for the network structures described in the previous section. For each training setup, we performed cross validation with the Leave-One-SuperTrial-Out (LOSO) setup of the JIGSAWS dataset \cite{JIGSAWS} (training on 4 super trials and leaving one super trial out for evaluation) and calculated the mean F1 scores across all folds. For each setup, we tuned the network parameters, including learning rate for the Adam Optimizer~\cite{kingma2014adam}, batch size and epoch number with nested cross validation. 

\subsubsection{Gesture Specific and Task Specific training (GSTS)}
In this setup separate models were trained for each gesture from each task, resulting in 10 different models. 
\subsubsection{Gesture Specific and Task Nonspecific training (GST*)}
This setup includes five separate models for different gesture classes (G1, G2, G3, G4, G6) by combining gesture-specific data from both tasks. 
\subsubsection{Gesture Nonspecific and Task Specific training (G*TS)}
In this setup two models were trained by combining all different gesture data for each task.

\subsubsection{Gesture Nonspecific and Task Nonspecific training (G*T*)}
This setup only included one model as the baseline by training on all the gesture data from both tasks. 

Although the total size of the dataset is fixed, the amount of data (both normal and erroneous examples) used for training the models in each setup is different. For example, the G*T* setup includes one model with the maximum training data size while the GSTS models are trained with smallest amount of data relevant to each specific gesture in each task. Table \ref{tb:gesture_specific} shows the total number of trials or gesture examples along with the number of erroneous gesture examples used for training in each setup. For evaluating the performance of different network architectures and comparing different training setups, we use the same testing setup and calculate the mean F1 score with cross validation. 




\subsection{Gesture Specific and Task Specific (\textbf{GSTS}) Models} 

\par In this experiment, we evaluated the performance of the CNN and LSTM networks versus the corresponding Siamese networks when trained with \textbf{GSTS} training. The results for different
\textbf{GSTS} models
are shown in the ``Suturing” and ``Needle Passing” rows in Table \ref{tb:gesture_specific}. Each cell in these two rows provides the mean F1 score for each 
\textbf{GSTS} model. 

\textbf{Observation 1: There are no apparent advantages of Siamese networks with \textbf{GSTS} training.} The Siamese network performance compared with that of its corresponding CNN or LSTM network varies across different gestures. For example, in G1 in task Suturing, CNN performs better than Siamese-CNN, and Siamese-LSTM performs better than LSTM. However, in G4 in task Suturing, Siamese-CNN performs better than CNN, and LSTM performs better than Siamese-LSTM. Hence, among \textbf{GSTS} models, there is no observed advantage of using Siamese networks over CNN and LSTM networks. This is likely due to model variance resulting from training with very small datasets.

\begin{figure}[t!]
\vspace{0.13cm}
    \centering
    \begin{subfigure}[t]{0.225\textwidth}
        \includegraphics[trim = 2mm 0mm 7mm 7mm, clip,width=\textwidth]{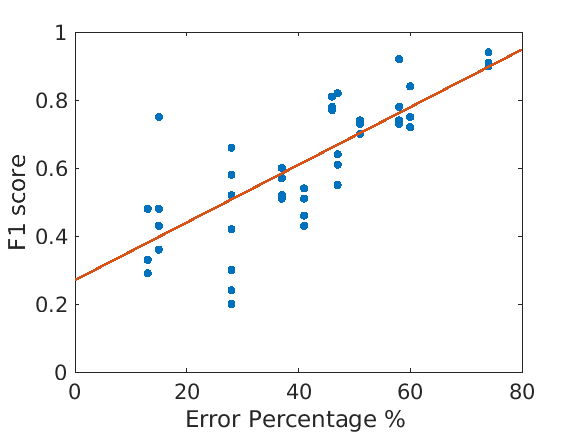}
    \vspace{-1.5em}
    \subcaption{\centering}
    \label{fig:F1_err}
    \end{subfigure}
    \begin{subfigure}[t]{0.25\textwidth}
        \includegraphics[clip,trim = {2cm 13.2cm 2.5cm 2.5cm}, width=\textwidth]{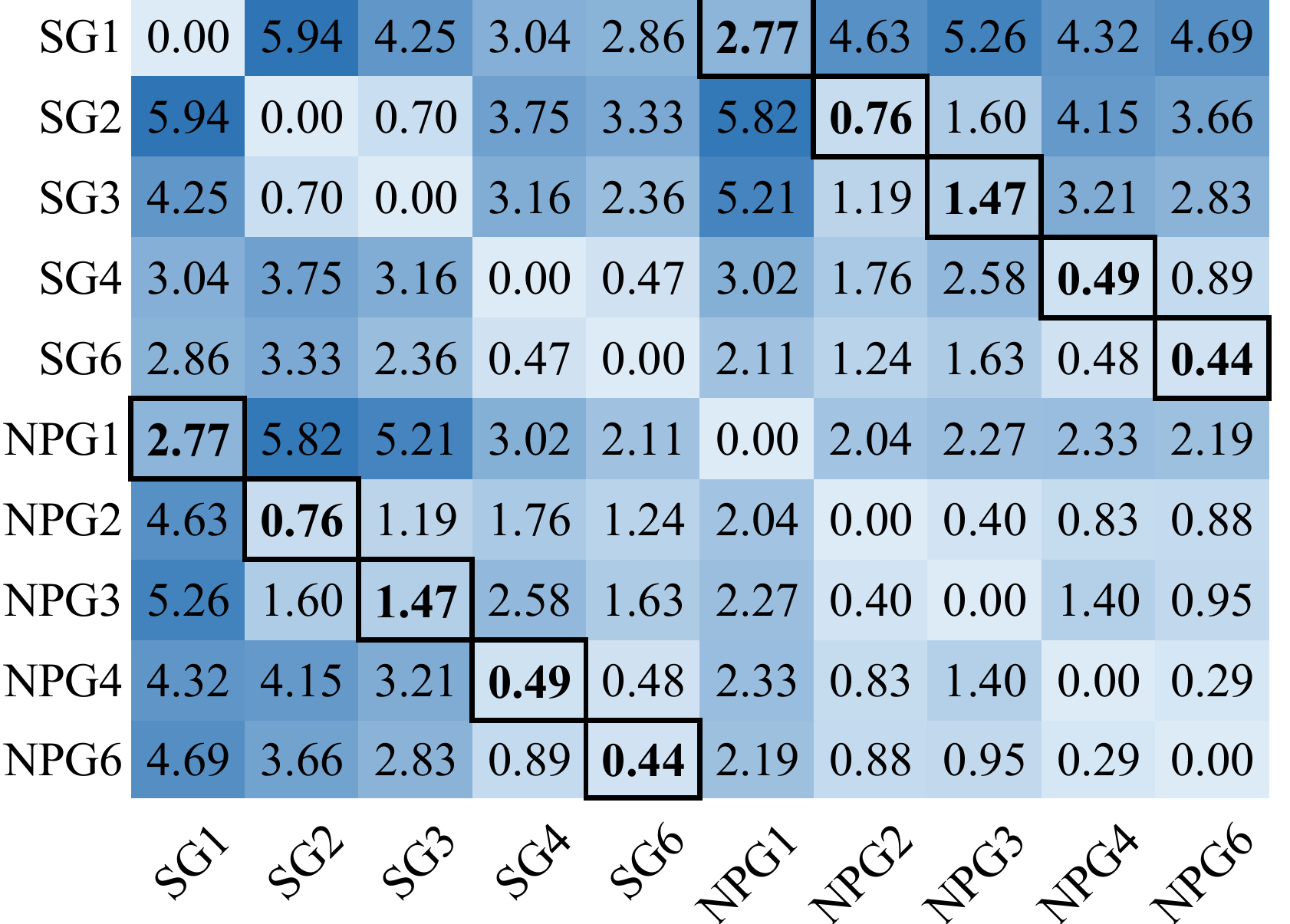}
        \vspace{-1.5em}
        \subcaption{\centering}
        \label{fig:KLD}
    \end{subfigure}
    \vspace{-0.5em}
    \caption{(a) F1 Scores vs. Error Percentages in Training Datasets, (b) Average KL Divergence Among Distribution of Normal Gestures}
    \vspace{-2em}
\end{figure}

\textbf{Observation 2: There is a positive correlation between error percentages and F1 scores for \textbf{GSTS} models.} Figure \ref{fig:F1_err} shows the percentage of erroneous trials in each dataset used for training each \textbf{GSTS} model versus the F1 score achieved in both the Suturing and Needle Passing tasks and for the four different network structures. We see a positive correlation between the F1 scores and error percentages across all the network structures. The same observation about the impact of error sample sizes on detection accuracy was made in \cite{Yasar_Alemzadeh_2020}. 

    \noindent\vsepfbox{
    \parbox{0.95\linewidth}{
    \textbf{Insight 1:} Having a higher percentage of error samples helps improve the performance of \textbf{GSTS} models.}
    }

\subsection{Gesture Specific and Task Nonspecific (\textbf{GST*})  Models}
In this experiment, our goal is to understand whether training \textbf{GST*} models that incorporate gesture data from different tasks can achieve better error detection performance than the \textbf{GSTS} models. This is motivated by the observation that the same gestures from different tasks share similar trajectories and error modes \cite{hutchinson2021analysis} and can provide more examples of errors of the same type for training. Thus, we combined the gesture data from the Suturing and Needle Passing tasks. We also compared the Siamese networks' performance with that of the CNN and LSTM networks. The row named ``Both Tasks" in Table \ref{tb:gesture_specific} shows the results for the \textbf{GST*} models. 

\textbf{Observation 1: \textbf{GST*} models achieve comparable or better performance than GSTS models for most cases in the four different network structures.} For the CNN network, G1, G2, G3 and G4's F1 scores as shown in ``Both Tasks" are greater than those from ``Suturing" and ``Needle Passing." For G6, the \textbf{GST*} model achieves comparable results to ``Suturing" (0.86 vs. 0.90), but better results than ``Needle Passing" (0.86 vs. 0.43). For the LSTM network, the F1 scores in ``Both Tasks" are comparable or greater than those for G1, G2, and G3 in Suturing and for G1, G3, G4, and G6 in Needle Passing. The F1 scores of the Siamese-CNN and Siamese-LSTM in ``Both Tasks" are greater than those in ``Suturing" and ``Needle Passing" for all gestures. 
\begin{figure*}[ht!]
\begin{subfigure}[b]{0.33\textwidth}
    \centering
    \includegraphics[width=\textwidth]{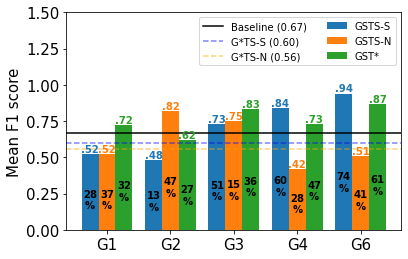}
    \vspace{-2em}
    \caption{LSTM}
    \label{fig:LSTM_NON}
\end{subfigure}
\begin{subfigure}[b]{0.33\textwidth}
    \centering
    \includegraphics[width=\textwidth]{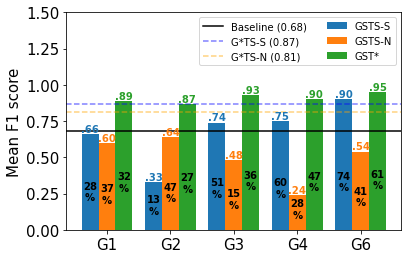}
    \vspace{-2em}
    \caption{Siamese LSTM}
    \label{fig:SLSTM_NON}
\end{subfigure}
\begin{subfigure}[b]{0.33\textwidth}
    \centering
    \includegraphics[width=\textwidth]{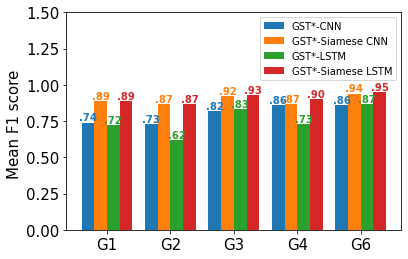}
    \vspace{-2em}
    \caption{Comparison between Different Networks}
    \label{fig:all_comp}
\end{subfigure}
\vspace{-1.5em}
\caption{Performance Comparison Across Different Networks and Training Setups. The values on top of each bar represent the F1 scores, and the percentage values at the middle of each bar represent the percentage of errors in the training data. }
\vspace{-1.75em}
\end{figure*}

\textbf{Observation 2: The baseline \textbf{G*T*} model can achieve better performance than \textbf{GSTS} models, but \textbf{GST*} models achieve better performance than the baseline.} 
The performance of the baseline classifier trained with all the data without the knowledge of gestures and tasks is shown in Figures \ref{fig:LSTM_NON} and \ref{fig:SLSTM_NON} as black lines. For the LSTM network, we observe that the baseline has better performance than the \textbf{GSTS} models for G1 and G2 in the Suturing task, and G1, G4, and G6 in the Needle Passing task. For the Siamese-LSTM network, the baseline has better performance in G1 and G2 for the Suturing task, and for all gestures in the Needle Passing task. This increase in performance is likely due to an increase in the number of training samples. Because the network sees more trajectory examples, it learns to better differentiate normal and erroneous trajectories. From Figures \ref{fig:LSTM_NON} and \ref{fig:SLSTM_NON}, we also note that the performance of the baseline is better than the \textbf{G*TS-S} and \textbf{G*TS-N} models shown as dashed lines for the LSTM network, but for the Siamese-LSTM network the baseline performance is worse. This could be because training with more varied data confuses the Siamese network, since it calculates dual input differences and performs voting to determine the final result. We will focus on the comparison between \textbf{GSTS}, \textbf{GST*} and \textbf{G*TS} in Section \ref{sec:G*TS}.
However, the \textbf{GST*} models (shown in  Figures \ref{fig:LSTM_NON} and \ref{fig:SLSTM_NON} with the green bars) still perform better than the baseline or the \textbf{G*TS} models  in most cases. When we combine gestures from both tasks, we increase the number of gesture specific  training samples.  
In order to investigate the similarity of normal gestures between tasks, we used KL divergence to quantify the difference in the distributions of the kinematic variables before data pre-processing.
In Figure \ref{fig:KLD}, the outlined black squares show the pairwise KL-divergence between distributions of the same gestures from two different tasks. These values are the smallest in the rows for G1, G4, and G6. Also in G2 and G3, 0.76 and 1.47 are smaller than most other elements in the row. 
A similar pattern holds when comparing Needle Passing to Suturing, but Needle Passing gestures are more similar to other Needle Passing gestures of different types. 
This shows that the gestures from Suturing are more similar to the same gestures in Needle passing than gestures of different types in the two tasks. 

\noindent\vsepfbox{
\parbox{0.95\linewidth}{
\textbf{Insight 2:} Increasing the amount of training data by adding \textit{more relevant data} (e.g., data from the same gesture class) improves the performance more than adding more irrelevant data.}  
    }

\textbf{Observation 3: \textbf{GST*} models can achieve better results than \textbf{GSTS} models and the baseline \textbf{G*T*} model despite having lower error percentages.}
From the insights of the \textbf{GSTS} training in the previous section, we noted that higher error percentage correlated with better performance. However, we observe that the \textbf{GST*} LSTM obtains a higher F1 score for G1 (0.72) with an error percentage of 32\% compared to the \textbf{GSTS} LSTM models for Suturing (0.52, 28\%) and Needle Passing (0.52, 37\%). This is also true for G3 where the \textbf{GST*} LSTM obtains an F1 score of 0.83 with an error percentage of 36\% and outperforms the \textbf{GSTS} LSTM models for Suturing (0.73, 51\%) and Needle Passing (0.75, 15\%). 
In Figure \ref{fig:SLSTM_NON} the Siamese-LSTM network also has higher F1 scores and lower error percentages for the \textbf{GST*} models for all gestures. We also calculated the error percentages for the \textbf{G*T*} model (baseline), which is 43\% for the black lines in both figures. We observe that \textbf{GST*} models (green bars in Figure \ref{fig:LSTM_NON} and Figure \ref{fig:SLSTM_NON}) sometimes achieve better performance than the baseline despite being trained on smaller datasets with lower error percentages (e.g., G1 and G3 in Figure \ref{fig:LSTM_NON} ; G1, G2 and G3 in Figure \ref{fig:SLSTM_NON}). 

\noindent\vsepfbox{
    \parbox{0.95\linewidth}{
    \textbf{Insight 3:} 
    Increasing the amount of relevant training data from the same gesture class improves the performance more than adding more error examples.}
    }

\textbf{Observation 4: Dual input Siamese networks perform better than their corresponding single input network structures with \textbf{GST*} training.} We compared the performance among the four networks with \textbf{GST*} training. In Figure \ref{fig:all_comp}, we observe that the average F1 scores for the Siamese networks are higher than their corresponding single networks. Siamese networks may perform better because of their double input structure that enables comparing and learning the difference between the two inputs. In addition, since the Siamese networks use pairs of inputs, the number of combinations of samples used for training is much greater than that of the single CNN or LSTM networks. 
\vspace{-1.0em}
\subsection{Gesture Nonspecific and Task Specific \textbf{G*TS} Models}
\label{sec:G*TS}
In this experiment, we evaluated the performance of all the network architectures when trained using the gesture nonspecific and task specific (\textbf{G*TS}) setup and compared them to the \textbf{GSTS} and \textbf{GST*} models. 
To compare the overall performance of different training setups, we evaluated the error detection performance by calculating the micro average F1 scores which provide a general performance metric and do not consider the individual performance for specific gestures. 

\textbf{Observation 1: \textbf{G*TS} training improves performance compared to \textbf{GSTS} training. However, \textbf{GST*} models have better performance than \textbf{G*TS} and \textbf{GSTS} models even though the training dataset for \textbf{G*TS} training is larger.} In Table \ref{tb:gesture_specific}, \textbf{G*TS} training has a higher micro F1 score than \textbf{GSTS} training for the Suturing task except for the CNN network. Similarly, for the Needle Passing task, \textbf{G*TS} training has a higher micro F1 score than \textbf{GSTS} except for the LSTM network. This is most likely due to an increase in the number of training samples when all gestures are combined to train a single model for each task. However, \textbf{GST*} training leads to the best micro F1 score compared to \textbf{G*TS} and \textbf{GSTS} training for all networks. 
In this case, the amount of data for \textbf{GST*} training is smaller than the \textbf{G*TS} training as shown in the ``Data" row in Table \ref{tb:gesture_specific}. This could be because increasing the relevant data from the same gesture class helps improve the network performance in comparison with just increasing the amount of training data (similar to \textbf{Insight 2} from the previous experiment). 

\textbf{Observation 2: Dual input Siamese networks perform better than single networks with \textbf{G*TS} training.} 
 As shown in Table \ref{tb:gesture_specific}, the Siamese-CNN has a greater micro F1 score than the CNN network for Suturing (0.79 vs. 0.72) and Needle Passing (0.74 vs. 0.56). The Siamese-LSTM also has a greater micro F1 score than the LSTM for Suturing (0.87 vs. 0.60) and Needle Passing (0.81 vs. 0.56).

  \noindent\vsepfbox{
  \parbox{0.95\linewidth}{
    \textbf{Insight 4}: Dual input Siamese networks achieve better performance likely due to their dual input structure that enables comparing normal/erroneous and normal/normal pairs of trajectories and learning models that better differentiate erroneous trajectories}
 }

\section{Discussion and Conclusion}
This paper evaluated the performance of dual input Siamese networks versus single CNN and LSTM network architectures for the task of erroneous gesture detection. We compared different training setups to assess how various levels of contextual knowledge on gestures and tasks affect the performance of each network. Our results show that the dual Siamese networks achieve higher accuracy than their corresponding single CNN and LSTM networks in most cases using the \textbf{GST*} and \textbf{G*TS} training setups. 
\par On the other hand, Siamese networks suffer from higher computational complexity than single networks. The single CNN networks take less computation time than LSTM networks across different training setups and gesture classes (avg. 0.50 ms vs. 1.26 ms). For both the Siamese-CNN and Siamese-LSTM networks, the computation time per window was proportional to the size of training data. This is because the Siamese networks pair each unknown window with all the known windows from the training data for voting and, thus, process more data at runtime. Also, Siamese-LSTM had much larger avg. computation time compared to Siamese-CNN across training setups and gestures (avg. 14.95 ms vs. 1.86 ms). This is because of hundreds of hidden states within the three LSTM layers that each must be computed sequentially. However, for the Siamese-CNN, outputs from the convolutional layers can be calculated in parallel on the GPU. Thus, for runtime error detection, using Siamese-CNN networks might be preferred and the parallel implementation of the Siamese models on GPU accelerators is required. 
\par Our results also show
that despite having fewer training samples, gesture specific models achieve better accuracy than generalized gesture nonspecific models. This further motivates the importance of developing models and labeled datasets that focus on gestures as the building blocks of procedures, their distributional similarities, and their unique error modes \cite{hutchinson2021analysis}. However, for runtime error detection using a gesture specific model, a gesture classifier is needed to first determine the gesture class for the input window~\cite{van2021gesture}. 
The extra computation time needed for gesture classification might negatively impact the performance and timeliness of error detection~\cite{Yasar_Alemzadeh_2020}. In addition, the gesture classifier might misclassify a given gesture, leading to the execution of an error detector that does not correspond to the gesture. 
\par Future work will focus on a more in-depth evaluation of the performance of the end-to-end gesture-specific error detection models, including the integration of the gesture classification and erroneous gesture detection models and analyzing the trade-offs between timeliness and accuracy.


\bibliographystyle{IEEEtran}
\bibliography{icrabib}

\end{document}